\documentclass[pmlr,twocolumn,10pt]{jmlr} 

\usepackage{booktabs}

\usepackage{siunitx}

\usepackage[switch]{lineno}
\usepackage{float}

\theorembodyfont{\upshape}
\theoremheaderfont{\scshape}
\theorempostheader{:}
\theoremsep{\newline}

\jmlrworkshop{Machine Learning for Health (ML4H) 2025} 

\title[Rethinking Tokenization for Clinical Time Series]{Rethinking Tokenization for Clinical Time Series: When Less is More}

\author{%
\Name{Rafi Al Attrach}\thanks{Shared first author.}\Email{rafiaa@mit.edu}\\
\addr Massachusetts Institute of Technology (MIT), USA; Technical University of Munich (TUM), Germany
\AND
\Name{Rajna Fani}\footnotemark[1]\Email{rajnaf@mit.edu}\\
\addr Massachusetts Institute of Technology (MIT), USA; Technical University of Munich (TUM), Germany
\AND
\Name{David Restrepo}\Email{david.restrepo@centralesupelec.fr}\\
\addr MICS, CentraleSupélec – Université Paris-Saclay, France
\AND
\Name{Yugang Jia}\Email{yugang@mit.edu}\\
\addr Massachusetts Institute of Technology (MIT), USA
\AND
\Name{Leo Anthony Celi}\thanks{Shared corresponding author.}\Email{lceli@mit.edu}\\
\addr Massachusetts Institute of Technology (MIT), USA; Harvard Medical School, USA; Beth Israel Deaconess Medical Center, USA
\AND
\Name{Peter Schüffler}\footnotemark[2]\Email{peter.schueffler@tum.de}\\
\addr Institute of Pathology, Technical University of Munich, Germany; Munich Center for Machine Learning (MCML), Germany 
}

\begin{document}

\maketitle

\begin{abstract}
Tokenization strategies shape how models process electronic health records, yet fair comparisons of their effectiveness remain limited. We present a systematic evaluation of tokenization approaches for clinical time series modeling using transformer-based architectures, revealing task-dependent and sometimes counterintuitive findings about temporal and value feature importance. Through controlled ablations across four clinical prediction tasks on MIMIC-IV, we demonstrate that explicit time encodings provide no consistent statistically significant benefit for the evaluated downstream tasks. Value features show task-dependent importance, affecting mortality prediction but not readmission, suggesting code sequences alone can carry sufficient predictive signal. We further show that frozen pretrained code encoders dramatically outperform their trainable counterparts while requiring dramatically fewer parameters. Larger clinical encoders provide consistent improvements across tasks, benefiting from frozen embeddings that eliminate computational overhead. Our controlled evaluation enables fairer tokenization comparisons and demonstrates that simpler, parameter-efficient approaches can, in many cases, achieve strong performance, though the optimal tokenization strategy remains task-dependent.
\end{abstract}

\begin{keywords}
tokenization, clinical time series, electronic health records, transformer models, healthcare AI
\end{keywords}

\paragraph*{Data and Code Availability}
We use the publicly available MIMIC-IV dataset~\citep{johnson2023mimiciv}, processed into the Medical Event Data Standard (MEDS) format~\citep{mcdermott2024meds} using the MIMIC-IV MEDS ETL pipeline~\citep{mimic_iv_meds_etl}. Our experiments are conducted using the MEDS-Torch framework~\citep{mcdermott2024medstorch}. Code for reproducing our experiments is available at \href{https://github.com/rafiattrach/rethinking-ehr-tokenization}{https://github.com/rafiattrach/rethinking-ehr-tokenization}.

\paragraph*{Institutional Review Board (IRB)}
This research uses the publicly available MIMIC-IV dataset, which has been previously approved for research use and does not require additional IRB approval for this analysis.

\section{Introduction}
\label{sec:intro}

Foundation models have transformed clinical machine learning~\citep{li2020behrt,rasmy2021medbert}. Tokenization, the process of converting raw electronic health record (EHR) events into dense vector representations, is a fundamental step that affects how models process information~\citep{ali2024tokenizer}. However, a key question persists: how should we tokenize complex, irregular medical time series data for optimal model performance? Unlike natural language or images, clinical data presents unique challenges: sparse measurements, irregular sampling, and heterogeneous feature types that demand specialized tokenization strategies. Recent benchmarks like EHRSHOT~\citep{wornow2023ehrshot} have highlighted the importance of systematic evaluation frameworks for clinical AI models, while generative approaches~\citep{waxler2025generative,pang2024cehrgpt} and foundation model applications~\citep{renc2025foundation} continue to expand the scope of clinical AI.

Recent frameworks like MEDS-Torch \citep{mcdermott2024medstorch} have proposed multiple tokenization approaches, with Triplet (structured code-time-value triplets) and TextCode (natural language descriptions with structured features) representing two different philosophies. The Triplet approach, building on STraTS~\citep{tipirneni2022strats}, represents each event as the sum of a learned code embedding, a time embedding, and a value embedding. In contrast, TextCode uses pretrained language models to encode code descriptions into semantic embeddings. While Triplet achieved the highest win rate and TextCode the lowest in initial benchmarks, these comparisons were potentially confounded by experimental design choices.

Despite recent innovations, establishing fair comparisons between tokenization strategies remains challenging. Prior benchmarks reported superior performance of Triplet-based representations over TextCode \citep{mcdermott2024medstorch}; however, these evaluations confounded several factors, including incomplete code-to-description mappings, the reliance on a single model size, and whether pretrained language encoders were frozen or fine-tuned. Such methodological inconsistencies make it difficult to attribute observed performance to the tokenization strategy itself rather than to experimental design choices.

To address these limitations, we conduct a series of systematic and controlled experiments designed to isolate the contribution of individual components. Our results challenge prevailing assumptions about clinical time series modeling, demonstrating that many widely adopted tokenization pipelines may be unnecessarily complex for certain tasks. By disentangling the effects of time encoding, value features, and embedding strategies, we enable clearer comparisons between tokenization approaches and highlight opportunities for simpler, more parameter-efficient models that maintain competitive performance.

\section{Methods}
\label{sec:methods}

\subsection{Clinical Tasks and Dataset}
Our evaluation uses four clinical prediction tasks from the MEDS-ACES task suite~\citep{xu2024aces}: in-hospital mortality within 24 hours of admission, ICU mortality within 24 hours of ICU stay, 1-year post-discharge mortality, and 30-day readmission. We use MIMIC-IV~\citep{johnson2023mimiciv}, processed into the Medical Event Data Standard (MEDS) format~\citep{mcdermott2024meds} using the MIMIC-IV MEDS ETL pipeline~\citep{mimic_iv_meds_etl}. This standardized format enables reproducible comparisons across different tokenization approaches.

\subsection{Tokenization Approaches}
We compare two tokenization strategies that represent different philosophies for encoding clinical events. \textbf{Triplet tokenization} represents each clinical event as the sum of three embeddings:
\begin{equation}
\mathbf{e}_{\text{triplet}} = \mathbf{W}_{\text{code}} \mathbf{c} + \text{CVE}_{\text{time}}(\mathbf{t}) + \text{CVE}_{\text{value}}(\mathbf{v})
\end{equation}
where $\mathbf{W}_{\text{code}}$ is a learned embedding matrix for medical codes $\mathbf{c}$, and CVE (Continuous Value Embedding)~\citep{tipirneni2022strats} maps scalar time deltas $\mathbf{t}$ and numeric values $\mathbf{v}$ to dense representations.

\begin{figure*}[!t]
\floatconts
  {fig:waterfall}
  {\caption{Triplet component ablation across four tasks. Error bars show $\pm$sd over $N=10$ seeds. Full numeric results are deferred to Appendix \tableref{tab:triplet_full}.}}
  {\centering\includegraphics[width=0.8\linewidth]{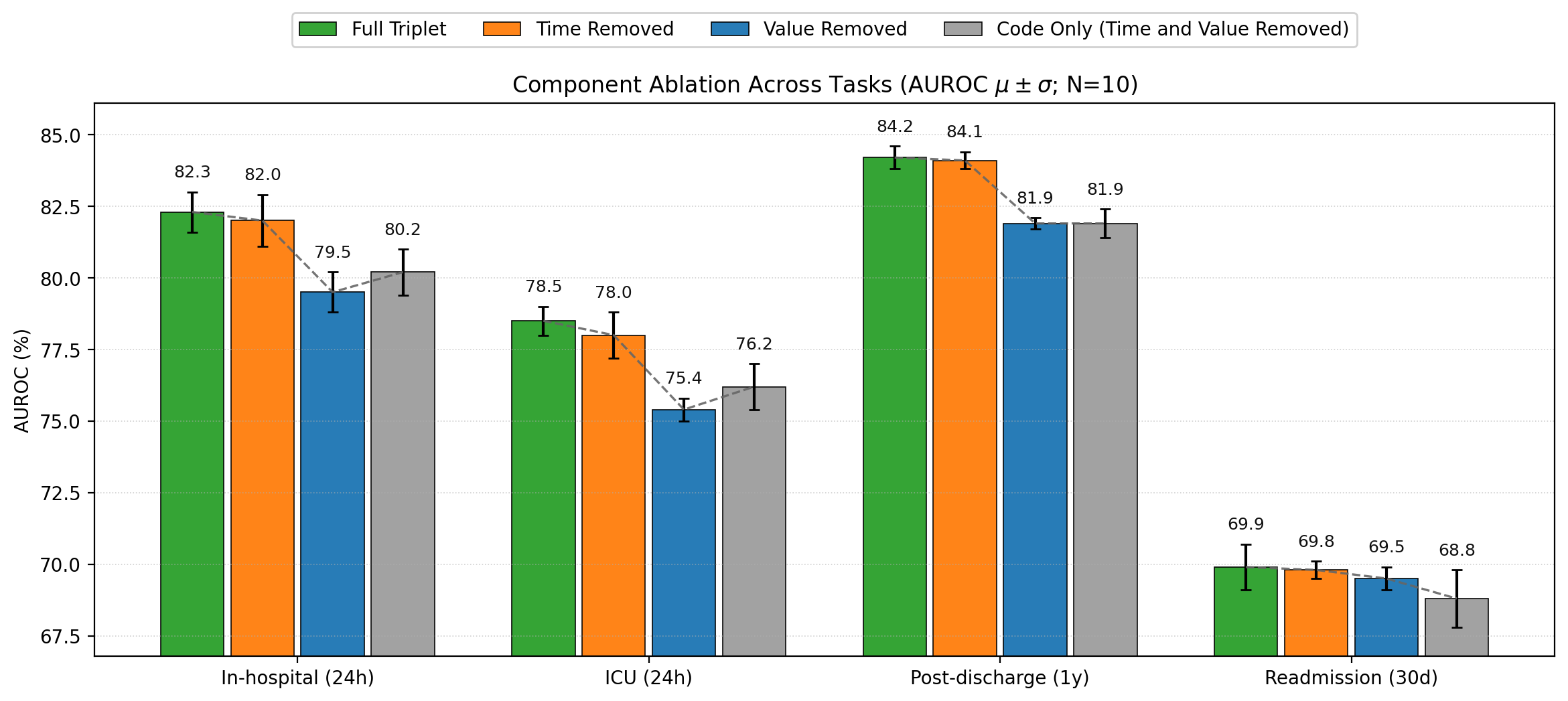}}
\end{figure*}

\textbf{TextCode} replaces the learned code embedding with representations from pretrained language models:
\begin{equation}
\mathbf{e}_{\text{textcode}} = \text{LM}(\text{desc}(\mathbf{c})) + \text{CVE}_{\text{time}}(\mathbf{t}) + \text{CVE}_{\text{value}}(\mathbf{v})
\end{equation}
where $\text{desc}(\cdot)$ maps medical codes to natural language descriptions and $\text{LM}$ represents a pretrained language model encoder.

\subsection{Controlled Experimental Design}
To enable fair comparisons, we systematically vary TextCode along four controlled axes: 1) mapping coverage, 2) training approach, 3) encoder scale, and 4) encoder domain. For mapping coverage, we compare original sparse mappings (25\% coverage) against enhanced mappings (100\% coverage) that we filled using the MIMIC-IV MCP Server~\citep{attrach2025conversational}. Here, coverage denotes the fraction of vocabulary codes with human‑readable descriptions (rather than fallback identifiers); the original mapping leaves most codes without descriptions, whereas the enhanced mapping provides descriptions for all codes (see Appendix, \sectionref{sec:enhanced_mapping}). This is critical because TextCode relies on natural language descriptions as input to language models~\citep{su2025multimodal, hegselmann2025large}. The training approach varies between trainable and frozen language model encoders. For encoder scale, we test models ranging from 15M to nearly 600M parameters. For encoder domain, we contrast clinical versus general-domain code encoders at comparable scale (both frozen) to isolate domain effects.

We conduct comprehensive component ablations on Triplet tokenization to understand the individual contribution of each modality. By systematically removing time features, value features, or both components while preserving the code information, we isolate the predictive signal contributed by each tokenization component across all clinical tasks.

\subsection{Statistical Analysis}
All models use supervised training in the standardized MEDS-Torch pipeline with transformer encoders. We report AUROC (Area Under the Receiver Operating Characteristic curve) as the primary metric~\citep{mcdermott2024closer}, aggregated over 10 random seeds. Statistical significance testing uses paired, two-sided Wilcoxon signed-rank tests, chosen for their robustness to non-normal distributions common in clinical prediction metrics. Bonferroni correction adjusts for multiple comparisons within each experimental group, with seed intersections used for proper pairing.

\section{Results}
\label{sec:results}

\subsection{Triplet component importance}
\textbf{A central finding is the limited impact of explicit time and value features for the evaluated tasks} (as shown in \figureref{fig:waterfall}). Removing time features shows no statistically significant effect across all clinical tasks tested, suggesting that explicit temporal encoding may provide less benefit than commonly assumed for these prediction tasks with transformer architectures. Remarkably, for readmission prediction, even removing value features shows no significant impact, indicating that event code sequences alone carry sufficient predictive signal. While value features do contribute to mortality prediction tasks, the consistent lack of temporal signal across all evaluated tasks reveals potential optimization opportunities in current tokenization approaches. Interestingly, for in-hospital and ICU mortality, the code-only model (with both time and value removed) outperformed the no-value variant, suggesting that isolated time features may introduce noise when value information is absent.

\subsection{TextCode: mapping, freezing, and code encoders}

\begin{table*}[!t]
\floatconts
  {tab:textcode}
  {\caption{TextCode results on post-discharge mortality. BioC-L is BioClinical-ModernBERT-L. $\dagger$ Bonferroni-corrected; * uncorrected $p<0.05$. Blank cells are not significant.}}
  {\centering\small
  \begin{tabular}{lcc}
  \toprule
  \textbf{Axis} & \textbf{Baseline} & \textbf{Variant} \\
  \midrule
  Mapping      & Original mapping: 71.2$\pm$0.8 & Enhanced mapping: 71.1$\pm$0.3 \\
  Trainability & Tiny-ClinicalBERT (Train.): 71.1$\pm$0.3 & Tiny-ClinicalBERT (Frozen): 81.1$\pm$1.3$^{\dagger}\uparrow$ \\
  Size         & Tiny-ClinicalBERT (Frozen): 81.1$\pm$1.3 & BioC-L (Frozen): 82.9$\pm$0.9$^{\dagger}\uparrow$ \\
  Domain       & Clinical (BioC-L): 82.9$\pm$0.9 & General (Qwen3): 81.5$\pm$1.2$^{\dagger}\downarrow$ \\
  \bottomrule
  \end{tabular}}
\end{table*}

\textbf{Freezing the code encoder dramatically improves performance, and larger clinical code encoders improve further}; this performance gain was statistically significant across all four tasks evaluated. BioClinical-ModernBERT-large substantially outperforms the baseline Tiny-ClinicalBERT model, while general-domain encoders like Qwen3-Embedding-0.6B achieve competitive but not superior performance compared to clinical models. Enhanced mapping coverage shows minimal impact compared to these architectural choices. \tableref{tab:textcode} summarizes post-discharge mortality, which is representative of the trends observed across other tasks (see Appendix for full results).

\noindent\emph{Axis definitions.} Each axis isolates a single experimental change against a controlled baseline configuration to isolate its effect:
\begin{itemize}
    \item \textbf{Mapping}: Original vs. Enhanced description coverage (using trainable Tiny-ClinicalBERT).
    \item \textbf{Trainability}: Trainable vs. Frozen Tiny-ClinicalBERT code encoder (using Enhanced mapping).
    \item \textbf{Size}: Tiny-ClinicalBERT (\(\approx\)15M) vs. BioClinical-ModernBERT-L (\(\approx\)396M) (both frozen).
    \item \textbf{Domain}: Clinical (BioClinical-ModernBERT-L) vs. General (Qwen3, \(\approx\)596M) (both frozen).
\end{itemize}

\subsection{Learnable time encodings}
Given that simple time encoding seemed ineffective, we also tested more complex learnable time encoders including Time2Vec~\citep{kazemi2019time2vec} and LeTE~\citep{chen2025rethinking}, but \textbf{they do not surpass the simple linear time feature}. Details are in Appendix~\sectionref{sec:time_encoders}.

\begin{figure}[H]
\floatconts
  {fig:efficiency}
  {\caption{Efficiency frontier for TextCode variants on post-discharge mortality. Bubble size reflects trainable parameters. Freezing the encoder offers a superior performance/cost trade-off.}}
  {\centering\includegraphics[width=0.95\linewidth]{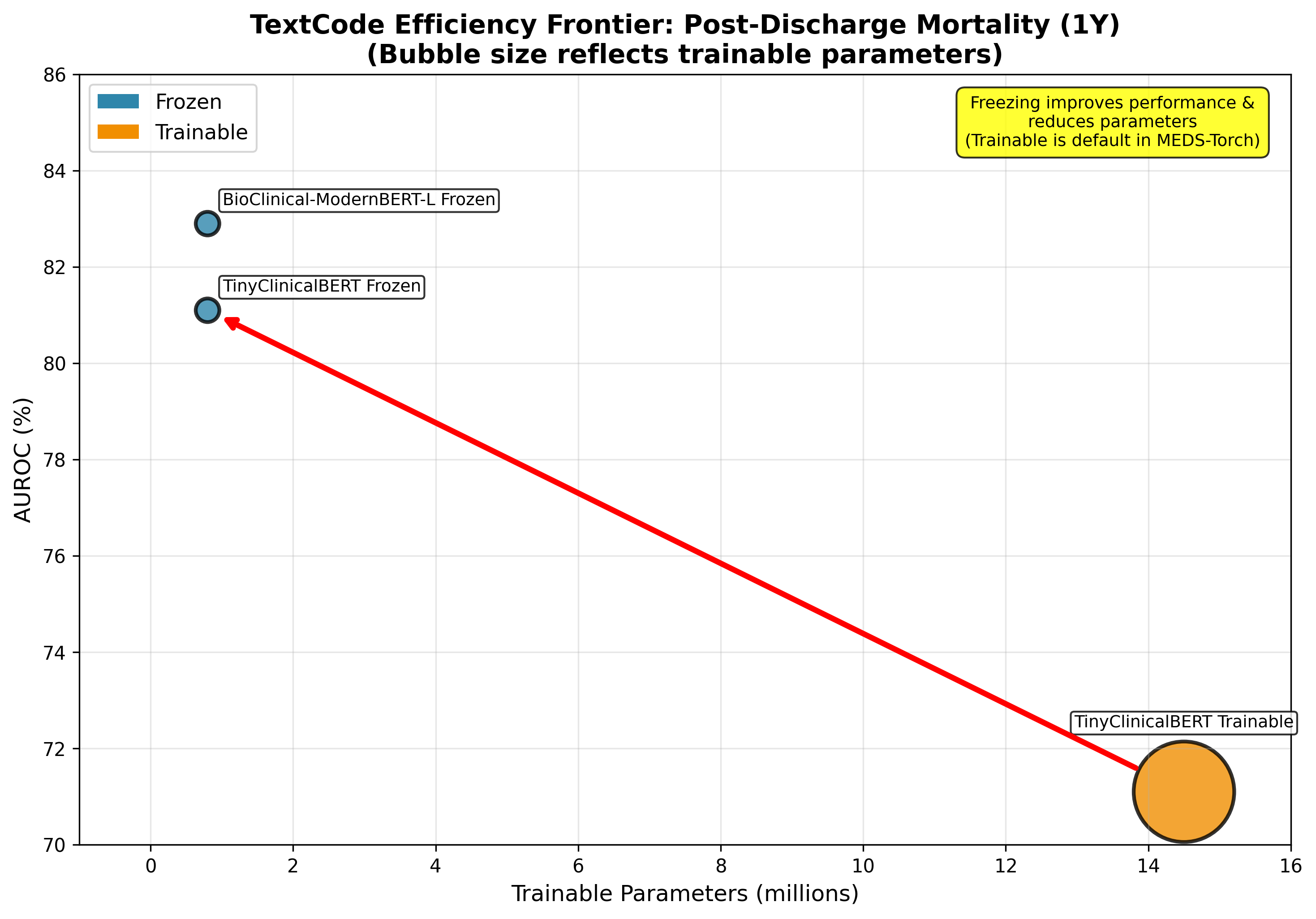}}
\end{figure}

\section{Conclusion}
\label{sec:conclusion}

\vspace{-1.0ex}

Our findings reveal that current EHR tokenization approaches may be over-engineered for certain clinical prediction tasks. The consistent lack of impact from explicit time features suggests that for transformer architectures applied to prediction tasks, these may provide limited additional signal beyond the inherent sequence order captured through positional embeddings. For mortality tasks, code-only models generally outperformed or matched models with code plus time embeddings, suggesting that time features without values may add noise rather than signal.

The strong performance of frozen encoders over trainable ones highlights a key trade-off. Fine-tuning entire LLMs risks catastrophic forgetting and overfitting, while requiring significant computational resources. In contrast, frozen encoders act as powerful, regularized feature extractors, leveraging rich pretrained knowledge while only training small projection layers. This approach proved more stable and effective across all evaluated tasks.

Our focus on Triplet and TextCode was motivated by their contrasting initial performance (highest vs lowest win rate in MEDS-Torch benchmarks) and their structural similarity that enables clean ablations. Given that code information emerged as the most critical predictive signal, we hypothesized that TextCode's underperformance reflected unfair experimental conditions rather than fundamental limitations. Our controlled comparison shows that frozen BioClinical-ModernBERT-Large substantially narrows TextCode's performance gap to Triplet, bringing it within approximately 1 AUROC point (compared to over 10 points in initial benchmarks), with Triplet maintaining a small advantage across tasks under our fixed experimental settings. EIC (Everything In Code), the third major tokenization paradigm in MEDS-Torch, discretizes all modalities into categorical tokens rather than using continuous encoders, requiring fundamentally different experimental designs beyond our current scope.

These findings suggest that simpler, more parameter-efficient approaches can achieve strong performance in clinical time series modeling, though optimal tokenization remains task-dependent. Our controlled evaluation framework enables fairer comparisons between tokenization methods. While these insights are derived from transformer-based models on MIMIC-IV for prediction tasks, they provide a foundation for broader investigation across architectures, datasets, and clinical applications. Future work should explore whether these patterns hold for other clinical tasks, model architectures (e.g., RNNs, state-space models) and across multiple institutions.

\bibliography{jmlr-sample}

\appendix

\section{Implementation Details}
\label{sec:implementation}

\subsection{TextCode Encoder Architecture}
\textbf{Freezing Mechanism:} The flexible TextCode encoder implements two distinct pathways. For frozen models, we precompute embeddings for all vocabulary codes using the pretrained language model, cache them as tensors, and disable gradient computation (\texttt{requires\_grad=False}). During training, embeddings are retrieved via fast tensor indexing rather than forward passes through the language model. For trainable models, we perform on-the-fly BERT inference with gradients enabled, allowing end-to-end optimization.

\textbf{Caching System:} Frozen models utilize a sophisticated caching mechanism that precomputes embeddings for the entire vocabulary ($\sim$1.2K codes) and stores them as pickle files. The cache maps vocabulary indices to native model dimensions (768 for Tiny-ClinicalBERT, 1024 for BioClinical-ModernBERT-large, 1024 for Qwen3), which are then projected to the token dimension (128). During training, unique codes are identified using \texttt{fast\_unique\_with\_inverse}, embeddings are retrieved via tensor indexing, and a learned projection layer maps to the model's token dimension (128).

\textbf{Enhanced Mapping:} \label{sec:enhanced_mapping} The original MEDS-Torch mapping suffered from 75\% missing descriptions, using fallback codes that may not capture clinical semantics. Our enhanced mapping transforms structured code identifiers into human-readable descriptions (e.g., ``INFUSION\_END//225166'' becomes ``Infusion of potassium chloride ended''), providing richer textual descriptions for language model encoding. Without rich textual descriptions, language model-based tokenizers may not distinguish between semantically related codes or leverage their extensive pretraining on medical literature. Our enhanced mapping ensures that any observed performance differences between Triplet and TextCode approaches can be attributed to the fundamental tokenization strategy rather than to data preprocessing artifacts.

\subsection{Model Specifications}
\begin{itemize}
    \item \textbf{Tiny-ClinicalBERT}~\citep{rohanian2023tinybert} (nlpie/tiny-clinicalbert): \(~15\)M parameters, 768-dim embeddings
    \item \textbf{BioClinical-ModernBERT-Large}~\citep{sounack2025bioclinicalmodernbert} (thomas-sounack/BioClinical-ModernBERT-large): 396M parameters, 1024-dim embeddings  
    \item \textbf{Qwen3-Embedding-0.6B}~\citep{zhang2025qwen3embedding} (Qwen/Qwen3-Embedding-0.6B): 596M parameters, 1024-dim embeddings
\end{itemize}

\subsection{Controlled Experimental Settings}
\begin{verbatim}
meds-torch-train experiment=triplet_mtr 
  paths.data_dir=triplet_tensors 
  paths.meds_cohort_dir=MEDS_cohort 
  data.task_root_dir=MEDS_cohort/tasks 
  trainer.accelerator=auto 
  trainer.devices=1 
  trainer.precision=32 
  trainer.strategy=auto 
  logger=csv 
  ++model.token_dim=128 
  ++data.dataloader.batch_size=64 
  ++trainer.max_epochs=10 
  ++data.dataloader.num_workers=6
\end{verbatim}

\subsection{Extended Time Encoder Results}
\label{sec:time_encoders}
\begin{itemize}
\item \textbf{Time2Vec:} We evaluated k $\in$ \{1, 2, 3, 4, 5, 6, 8, 10, 20, 50\} for the sinusoidal component count using a standard PyTorch implementation~\citep{garcia_time2vec}. Raw p-values showed occasional improvements (k=6,10 for in-hospital mortality AUROC), but none survived Bonferroni correction across the full experimental grid.
\item \textbf{LeTE:} Learnable time encodings mixed Fourier basis functions with spline multilayer perceptrons. We tested balanced (equal Fourier/spline weights), spline-heavy, and Fourier-heavy configurations. All variants performed comparably to the baseline continuous value encoder, suggesting that simple linear time encoding suffices for these clinical prediction tasks.
\end{itemize}

\subsection{Computational Efficiency}
Frozen TextCode models achieve 15× reduction in trainable parameters (14.5\,M $\to$ $<\!1$\,M) while improving performance. Training time decreases substantially due to cached embeddings eliminating repeated language model forward passes. Even with larger frozen models (396M ModernBERT-Large), the computational footprint during training remains minimal since only the projection layer requires gradient computation.

\section{Complete Ablation Results}
\label{sec:full_results}
Complete numeric results for the Triplet component ablations across all four clinical tasks are provided in \tableref{tab:triplet_full}.

\begin{table*}[!t]
\floatconts
  {tab:triplet_full}
  {\caption{Triplet ablations (AUROC $\mu\pm\sigma$; $N{=}10$). $^{\dagger}$ Bonferroni-corrected $p<0.05$; $^{*}$ uncorrected. Direction arrows compare to Triplet.}}
  {\centering\small
  \begin{tabular}{lcccc}
  \toprule
  \textbf{Task} & \textbf{Triplet} & \textbf{No time} & \textbf{No value} & \textbf{Code-only} \\
  \midrule
  In-hospital mortality (24h) & 82.3$\pm$0.7 & 82.0$\pm$0.9 & 79.5$\pm$0.7$^{\dagger}\downarrow$ & 80.2$\pm$0.8$^{\dagger}\downarrow$ \\
  ICU mortality (24h) & 78.5$\pm$0.5 & 78.0$\pm$0.8 & 75.4$\pm$0.4$^{\dagger}\downarrow$ & 76.2$\pm$0.8$^{\dagger}\downarrow$ \\
  Post-discharge mortality (1y) & 84.2$\pm$0.4 & 84.1$\pm$0.3 & 81.9$\pm$0.2$^{\dagger}\downarrow$ & 81.9$\pm$0.5$^{\dagger}\downarrow$ \\
  Readmission (30d) & 69.9$\pm$0.8 & 69.8$\pm$0.3 & 69.5$\pm$0.4 & 68.8$\pm$1.0$^{*}\downarrow$ \\
  \bottomrule
  \end{tabular}}
\end{table*}

\section{Full TextCode Results Across All Tasks}
\label{sec:textcode_full}
Full per-task results for the TextCode experiments are provided in \tableref{tab:textcode_full_table}.

\begin{table*}[!t]
\floatconts
  {tab:textcode_full_table}
  {\caption{Full TextCode results across all tasks (AUROC $\mu\pm\sigma$; $N{=}10$).}}
  {\centering\small
  \begin{tabular}{lcccc}
  \toprule
  \textbf{Variant} & \textbf{In-hosp. mort.} & \textbf{ICU mort.} & \textbf{Post-disch. mort.} & \textbf{Readmission} \\
  \midrule
  Tiny-ClinicalBERT (Train.) & 69.5$\pm$1.3 & 64.3$\pm$0.8 & 71.1$\pm$0.3 & 61.1$\pm$0.4 \\
  Tiny-ClinicalBERT (Frozen) & 72.5$\pm$1.6 & 72.8$\pm$0.9 & 81.1$\pm$1.3 & 65.6$\pm$1.1 \\
  BioClinical-ModernBERT-Large (Frozen)   & 76.2$\pm$1.9 & 75.6$\pm$1.3 & 82.9$\pm$0.9 & 69.5$\pm$1.7 \\
  Qwen3 (Frozen)    & 75.1$\pm$1.2 & 72.8$\pm$0.9 & 81.5$\pm$1.2 & 67.2$\pm$1.2 \\
  \bottomrule
  \end{tabular}}
\end{table*}

\section{Learnable Time Encoders}
\label{sec:learnable_time}

We evaluated Time2Vec and LeTE. For Time2Vec, we explored a range of values for the number of sinusoidal components, $k$. We began with small values and increased the step size after observing diminishing returns. No variant offered a statistically significant improvement over the simple linear baseline, as shown in \tableref{tab:time_full_table}.

\begin{table*}[!t]
\floatconts
  {tab:time_full_table}
  {\caption{Best learnable time encoding per task vs baseline (AUROC $\mu\pm\sigma$; \(N{=}10\)).}}
  {\centering\small
  \begin{tabular}{lccc}
  \toprule
  \textbf{Task} & \textbf{Baseline (CVE)} & \textbf{Best Variant} & \textbf{Wilcoxon $p$} \\
  \midrule
  In-hospital mortality (24h)   & 82.3$\pm$0.7 & T2V(k=6): 82.7$\pm$0.8 & 0.193 \\
  ICU mortality (24h)           & 78.5$\pm$0.5 & Spline (Heavy): 78.9$\pm$0.4        & 0.037 \\
  Post-discharge mortality (1y) & 84.2$\pm$0.4 & Spline (Bal.): 84.8$\pm$0.8     & 0.027 \\
  Readmission (30d)             & 69.9$\pm$0.8 & T2V(k=2): 70.3$\pm$0.7  & 0.037 \\
  \bottomrule
  \end{tabular}}
\end{table*}

\end{document}